\documentclass[11pt,a4paper]{article}
\usepackage{authblk}
\usepackage[hyperref]{acl2020}
\usepackage{times}
\usepackage{latexsym}
\usepackage{amsfonts}
\usepackage{longtable}
\usepackage{fontawesome}
\definecolor{lightgray2}{gray}{0.9}
\usepackage{graphicx}
\usepackage{multirow}
\graphicspath{ {./} }

\aclfinalcopy 

\title{CompRes: A Dataset for Narrative Structure in News}

\author[1]{Effi Levi}
\author[2]{Guy Mor}
\author[2]{Shaul Shenhav}
\author[2,3]{Tamir Sheafer}
\affil[1]{Institute of Computer Science, The Hebrew University}
\affil[2]{Department of Political Science, The Hebrew University}
\affil[3]{Department of Communication and Journalism, The Hebrew University}
\affil[ ]{{\tt efle@cs.huji.ac.il}} 
\affil[ ]{{\tt \{guy.mor$|$shaul.shenhav$|$tamir.sheafer\}@mail.huji.ac.il}}

\date{}

\begin{document}
\maketitle
\begin{abstract}

This paper addresses the task of automatically detecting narrative structures in raw texts. Previous works have utilized the oral narrative theory by Labov and Waletzky to identify various narrative elements in personal stories texts. Instead, we direct our focus to news articles, motivated by their growing social impact as well as their role in creating and shaping public opinion.

We introduce CompRes -- the first dataset for narrative structure in news media. We describe the process in which the dataset was constructed: first, we designed a new narrative annotation scheme, better suited for news media, by adapting elements from the narrative theory of Labov and Waletzky ({\tt Complication} and {\tt Resolution}) and adding a new narrative element of our own (\texttt{Success}); then, we used that scheme to annotate a set of 29 English news articles (containing 1,099 sentences) collected from news and partisan websites. We use the annotated dataset to train several supervised models to identify the different narrative elements, achieving an $F_1$ score of up to 0.7. We conclude by suggesting several promising directions for future work.

\end{abstract}

\section{Introduction}
\label{sec:intro}

Automatic extraction of narrative structures from texts is a multidisciplinary field of research, combining discourse and computational theories, which has been receiving increasing attention over the last few years. Examples include modeling narrative structures for story generation \cite{gervas2006narrative}, using unsupervised methods to detect narrative event chains \cite{chambers-jurafsky-2008-unsupervised} 
and detecting content zones \cite{baiamonte2016annotating} in news articles, using semantic features to detect \textit{narreme} boundaries in fictitious prose \cite{delmonte-marchesini-2017-semantically}, identifying turning points in movie plots \cite{papalampidi2019movie} and using temporal word embeddings to analyze the evolution of characters in the context of a narrative plot \cite{volpetti2020temporal}.

A recent and more specific line of work focuses on using the theory laid out in \citet{labov1967narrative} and later refined by \citet{labov2013language} to characterize narrative elements in personal experience texts. \citet{swanson-etal-2014-identifying} relied on \citet{labov1967narrative} to annotate a corpus of 50 personal stories from weblogs posts, and tested several models over hand-crafted features to classify clauses into three narrative clause types: \textit{orientation}, \textit{evaluation} and \textit{action}.
\citet{ouyang-mckeown-2014-towards} constructed a corpus from 20 oral narratives of personal experience collected by \citet{labov2013language}, and utilized logistic regression over hand-crafted features to detect instances of \textit{complicating actions}. 

While these works concentrated their effort on detecting narrative elements in personal experience texts, we direct our focus to detecting narrative structure in news stories; the social impact of news stories distributed by the media and their role in creating and shaping of public opinion incentivized our efforts to adapt narrative structure analysis to this domain. To the best of our knowledge, ours is the first attempt to automatically detect the narrative elements from \cite{labov2013language} in news articles.

In this work, we introduce CompRes -- a new dataset of news articles annotated with narrative structure. For this purpose, we adapted two elements from the narrative theory presented in \citet{labov1967narrative, labov1972language, labov2013language}, namely {\tt Complication} and {\tt Resolution}, while adding a new narrative element, {\tt Success}, to create a new narrative annotation scheme which is better suited for informational text rather than personal experience. We used this scheme to annotate a newly-constructed corpus of 29 English news articles, containing a total of 1099 sentences; each sentence was tagged with a subset of the three narrative elements (or, in some cases, none of them), thus defining a novel multi-label classification task. 

We employed two supervised models in order to solve this task; a baseline model which used a linear SVM classifier over a bag-of-words feature representation, and a complex deep-learning model -- a fine-tuned pre-trained state-of-the-art language model (RoBERTa-based transformer). The latter significantly outperformed the baseline model, achieving an average $F_1$ score of 0.7.

The remainder of this paper is organized as follows: Section \ref{sec:narrativeanalysis} gives a theoretical background and describes the adjustments we have made to the scheme in \cite{labov2013language} in order to adapt it to informational text. Section \ref{sec:dataset} provides a complete description of the new dataset and of the processes and methodologies which were used to construct and annotate it, along with a short analysis and some examples for annotated sentences. Section \ref{sec:experiments} describes the experiments conducted on the dataset, reports and discusses our preliminary results. Finally, Section \ref{sec:conclusion} contains a summary of our contributions as well as several suggested directions for future work.

\section{Narrative Analysis}
\label{sec:narrativeanalysis}

\subsection{Background}

The study of narratives has always been associated, in one way or another, with an interest in the structure of texts. Ever since the emergence of formalism and structuralistic literary criticism \cite{propp1968morphology}
and throughout the development of narratology \cite{genette1980narrative, fludernik2009introduction, chatman1978existents, rimmon2003narrative},
narrative structure has been the focus of extensive theoretical and empirical research. While most of these studies were conducted in the context of literary analysis, the interest in narrative structures has made inroads into social sciences. The classical work by \citet{labov1967narrative} on oral narratives, as well as later works \cite{labov1972language, labov2013language}, signify this stream of research by providing a schema for an overall structure of narratives, according to which a narrative construction encompasses the following building blocks  \cite{labov1972language, labov2013language}:
 \begin{itemize}
     \item \textit{abstract}, i.e. what the narrative is about
     \item \textit{orientation}, i.e. the time, the place and the persons 
     \item \textit{complicating action}, explained in Section \ref{subsec:adaptation}
     \item \textit{evaluation}, i.e. revealing the narrator’s attitude towards the narrative or the meaning given to the events
     \item \textit{resolution}, explained in Section \ref{subsec:adaptation}
     \item \textit{coda}, i.e. brings the time of reference back to the present time of narration
 \end{itemize}
These building blocks provide useful and influential guidelines for a structural analysis of oral narratives.         
\subsection{Adaptation}
\label{subsec:adaptation}

Despite the substantial influence of \cite{labov1967narrative, labov2013language}, scholars in the field of communication have noticed that this overall structure does not necessarily comply with the form of news stories \cite{thornborrow2004storying, bell1991language, van1988news}
and consequently proposed simpler narrative structures \cite{thornborrow2004storying}. In line with this stream of research, our coding scheme was highly attentive to the unique features of news articles. A special consideration was given to the variety of contents, forms and writing styles typical for media texts. For example, we required a coding scheme that would fit  laconic or problem-driven short reports (too short for full-fledged “Labovian” narrative style), as well as complicated texts with multiple story-lines moving from one story to another. We addressed this challenge by focusing on two out of Labov’s six elements - \textit{complicating action} and \textit{resolution}. Providing “answers to the potential question ‘And then what happened?’” \cite{labov2013language}, we consider these two elements to be the most fundamental and relevant for news analysis. There are several reasons for our focus on these particular elements: first, it goes in line with the understanding that worth-telling stories usually consist of protagonists facing and resolving problematic experiences \cite{eggins2005analysing}; from a macro-level perspective, this can be useful to capture or characterize the plot type of stories \cite{shenhav2015analyzing}. Moreover, these elements resonate with what is considered by \citet{entman2004projections} to be the most important Framing Functions - problem definition and remedy. Our focus can also open up opportunities for further exploration of other important narrative elements in media stories, such as identifying “villainous” protagonists who are expected to be strongly associated with the complication of the story, and who are expected to be instrumental to a successful resolution \cite{shenhav2015analyzing}.  
In order to adapt the original \textit{complicating action} and \textit{resolution} categories to news media content, we designed our annotation scheme as follows. Complicating action -- hence, \texttt{Complication} -- was defined in our narrative scheme as an event, or series of events, that point at problems or tensions. \texttt{Resolution} refers to the way the story is resolved or to the release of the tension. An improvement from -- or a manner of -- coping with an existing or hypothetical situation was also counted as a resolution. We did that to follow the lack of a “closure” which is typical for many social stories \cite{shenhav2015analyzing} and the often tentative or speculative notion of future resolutions in news stories \cite{thornborrow2004storying}. We have therefore included in this category any temporary or partial resolutions. The transitional characteristic of the resolution brought us to subdivide this category into yet another derivative category defined as \texttt{Success}. Unlike the transitional aspect of the resolution, which refers, implicitly or explicitly, to a prior situation, this category was designed to capture any description or indication of an achievement or a good and positive state. 
\section{The CompRes Dataset}
\label{sec:dataset}

Here we describe the process of constructing CompRes, our dataset of news articles annotated with narrative structures. The dataset contains 29 news articles, comprising 1,099 sentences. An overview of the dataset is given in Table \ref{tab:datasetoverview}.

\begin{table*}[t]
\centering
\begin{tabular}{lccc}
\hline & \textbf{Complication} & \textbf{Resolution} & \textbf{Success} \\ \hline
\# Sentences & 570 & 269 & 176 \\
Proportion in Dataset & 52\% & 24\% & 16\% \\
\hline
\end{tabular}
\caption{\label{tab:datasetoverview} Overview of the CompRes dataset. Note that the categories are not mutually exclusive, due to the multi-labeled nature of the annotation scheme}
\end{table*}

\subsection{Pilot Study}
\label{subsec:pilot}

We started by conducting a pilot study, for the purpose of formalizing an annotation scheme and training our annotators. For this study, samples were gathered from print news articles in the broad domain of economics, published between 1995 and 2017 and collected via {\it LexisNexis}. We used these articles to refine elements from the theory presented in \cite{labov1967narrative,labov2013language} into a narrative annotation scheme which is better suited for news media (as detailed in Section \ref{subsec:adaptation}), as well as perform extensive training for our annotators. The result was a multi-label annotation scheme containing three narrative elements: {\tt Complication}, {\tt Resolution} and {\tt Success}.

Following the conclusion of the pilot study, we used the samples which were collected and manually annotated during the pilot to train a multi-label classifier for this task by fine-tuning a RoBERTa-base transformer \cite{liu2019roberta}. This classifier was later used to provide labeled candidates for the annotators during the annotation stage of the CompRes dataset, in order to optimize annotation rate and accuracy. The pilot samples were then discarded.

\subsection{News Articles}

The news articles for the CompRes dataset were sampled from 120 leading news and partisan websites in the English language, all published between 2017 and 2020. The result is a corpus of 29 news articles comprising a total of 1,099 sentences, with an average of 39.3 sentences per article (and a standard deviation of 21.8), and an average of 22.2 tokens per sentence (with a standard deviation of 13.0). The articles are semantically diverse, as they were sampled from a wide array of topics such as politics, economy, sports, culture, health. For each article in the corpus, additional meta-data is included in the form of the article title and the URL from which the article was taken (for future reference).

\subsection{Preprocessing}

The news articles' content was extracted using \href{http://diffbot.com}{diffbot}. The texts were scraped and split into sentences using the Punkt unsupervised sentence segmenter \cite{kiss-strunk-2006-unsupervised}. Some remaining segmentation errors were manually corrected.

\subsection{Annotation}

\subsubsection{Guidelines}
\label{subsubsec:guidelines}

Following the pilot study (Section \ref{subsec:pilot}), a code book containing  annotation guidelines was produced. For each of the three categories in the annotation scheme -- {\tt Complication}, {\tt Resolution} and {\tt Success} -- the guidelines provide: 
\begin{itemize}
    \item A general explanation of the category
    \item A list of well-defined criteria for identifying the category
    \item Select examples of sentences labeled exclusively with the category
\end{itemize}

\subsubsection{Process}

We employed a three-annotator setup for annotating the collected news articles. First, the model which was trained during the pilot stage (Section \ref{subsec:pilot}) was used to produce annotation suggestions for each of the sentences in the corpus. Each sentence was then separately annotated by two trained annotators according to the guidelines described in Section \ref{subsubsec:guidelines}. Each annotator had the choice to either accept the suggested annotation or to change it by adding or removing any of the suggested labels. Disagreements were later decided by a third expert annotator (the project lead).

Table \ref{tab:intercoderreliability} reports inter-coder reliability scores for each of the three categories, averaged across pairs of annotators: the raw agreement (in percentage) between annotators, and Cohen's Kappa coefficient, accounting for chance agreement \cite{artstein-poesio-2008-survey}. 

\begin{table*}[t]
\centering
\begin{tabular}{lccc}
\hline & \textbf{Complication} & \textbf{Resolution} & \textbf{Success} \\ \hline
Avg. raw agreement & 90.9\% & 87.9\% & 91.4\% \\
Avg. Cohen's Kappa & 0.82 & 0.64 & 0.7\\
\hline
\end{tabular}
\caption{\label{tab:intercoderreliability} Inter-coder reliability}
\end{table*}

\subsection{Analysis}

Categories vary significantly in their prevalence in the corpus; their respective proportions in the dataset are given in Table \ref{tab:datasetoverview}. The categories are unevenly distributed: {\tt Complication} is significantly more frequent than {\tt Resolution} and {\tt Success}. This was to be expected, considering the known biases of "newsworthiness" towards problems, crises and scandals, and due to the fact that in news media, resolutions often follow reported complications.

Table \ref{tab:label-correlations-table} reports pairwise Pearson correlations ($\phi$ coefficient) between the categories. A minor negative correlation was found between {\tt Complication} and {\tt Success} ($\phi=-0.26$), and a minor positive correlation was found between {\tt Resolution} and {\tt Success} ($\phi=0.22$); these were not surprising, as success is often associated with resolving some complication. However, {\tt Complication} and {\tt Resolution} were found to be completely uncorrelated ($\phi=0.01$), which -- in our opinion -- indicates that the \texttt{Success} category does indeed bring added value to our narrative scheme.

\begin{table}
\centering
\begin{tabular}{lccc}
\hline & \textbf{Comp.} & \textbf{Res.} & \textbf{Suc.} \\ \hline
\textbf{Comp}. & 1 & & \\
\textbf{Res.} & 0.01 & 1 & \\
\textbf{Suc.} & -0.26 & 0.22 & 1 \\
\hline
\end{tabular}
\caption{\label{tab:label-correlations-table} Inter-category correlations ($\phi$ coefficient)}
\end{table}

In Table \ref{tab:examples} we display examples of annotated sentences from the CompRes dataset. Note that all the possible combinations of categories exist in the dataset; Table \ref{tab:category-combinations} summarizes the occurrences of each of the possible category combinations in the dataset.

The fact that the dataset is composed of full coherent news articles allows the analysis of a range of micro, meso and macro stories in narrative texts. For example, an article in the dataset concerning the recent coronavirus outbreak in South Korea\footnote{https://edition.cnn.com/2020/03/09/asia/south-korea-coronavirus-intl-hnk/index.html} opens with a one-sentence summary, tagged with both \texttt{Complication} and \texttt{Resolution}: 
\begin{itemize}
    \item[] "South Korea's top public health official hopes that the country has already gone through the worst of the novel coronavirus outbreak that has infected thousands inside the country." (\texttt{Complication}, \texttt{Resolution})
\end{itemize}

This problem-solution (or in this case, hopeful solution) plot structure reappears in the same article, but this time it is detailed over a series of sentences: 
\begin{itemize}
    \itemsep0em 
    \item[] “More than 7,300 coronavirus infections have been confirmed throughout South Korea, killing more than 50." (\texttt{Complication})
    \item[] "It is one of the largest outbreaks outside mainland China, where the deadly virus was first identified.” (\texttt{Complication})
    \item[] “However, the number of new daily infections in South Korea has declined in recent days.” (\texttt{Complication}, \texttt{Resolution})
    \item[] “… while he believes the aggregate number of infections is high, he is confident in the job South Korea did to combat the virus' spread and would advise other governments…” (\texttt{Complication}, \texttt{Resolution})
    \item[] “The South Korean government has been among the most ambitious when it comes to providing the public with free and easy testing options." (\texttt{Success})
\end{itemize}

The sequence starts with two sentences tagged with \texttt{Complication}, followed by two additional ones tagged with both  \texttt{Complication} and \texttt{Resolution}, and concludes with a sentence tagged as \texttt{Success}. This example demonstrates a more gradual transition from problem through solution to success.

\begin{table}
\centering
\begin{tabular}{lc}
\hline & \textbf{\# Sentences} \\ \hline
Complication \& Resolution & 143 \\
Complication \& Success & 39 \\
Resolution \& Success & 81 \\
All Three Categories & 28 \\
\hline
\end{tabular}
\caption{\label{tab:category-combinations} Occurrences of category combinations in the CompRes dataset}
\end{table}

\begin{table*}
    \centering
    \begin{tabular}{lp{11cm}|ccc}
       \hline \textbf{\#} & \textbf{Sentence} & \textbf{Comp.} & \textbf{Res.} & \textbf{Suc.}  \\ \hline
1 &	It is no surprise, then, that the sensational and unverified accusations published online this week stirred a media frenzy.	& \faCheck	& \faClose	& \faClose	\\
& & & & \\
2 &	America would lose access to military bases throughout Europe as well as NATO facilities, ports, airfields, etc.	& \faCheck	& \faClose	& \faClose	\\
& & & & \\
3 &	How did some of the biggest brands in care delivery lose this much money?	& \faCheck	& \faClose	& \faClose \\
& & & & \\
4 &	Bleeding from the eyes and ears is also possible after use, IDPH said.	& \faCheck	& \faClose	& \faClose	\\
& & & & \\
5 &	The gentrification project, which concluded this year, included closing more than 100 brothels and dozens of coffee shops (where cannabis can be bought), and trying to bring different kinds of businesses to the area.	& \faClose	& \faCheck	& \faClose	\\
& & & & \\
6 & His proposal to separate himself from his business would have him continue to own his company, with his sons in charge.	& \faClose	& \faCheck	& \faClose	\\
& & & & \\
7 & Instead, hospitals are pursuing strategies of market concentration.	& \faClose	& \faCheck	& \faClose	\\
& & & & \\
8 & The South Korean government has been among the most ambitious when it comes to providing the public with free and easy testing options. & \faClose	& \faClose	& \faCheck	\\
& & & & \\
9 &	The husband and wife team were revolutionary in this fast-changing industry called retail.	& \faClose	& \faClose	& \faCheck	\\
& & & & \\
10 &	With its centuries-old canals, vibrant historic center and flourishing art scene, Amsterdam takes pride in its cultural riches.	& \faClose	& \faClose	& \faCheck	\\
& & & & \\
11 &	Mr. Trump chose to run for president, he won and is about to assume office as the most powerful man in the world.	& \faClose	& \faClose	& \faCheck	\\
& & & & \\
12 &	Soon after, her administration announced a set of measures intended to curb misconduct.	& \faCheck	& \faCheck	& \faClose	\\
& & & & \\
13 &	Voter suppression is an all-American problem we can fight - and win.	& \faCheck	& \faCheck	& \faClose	\\
& & & & \\
14 &	Though many of his rivals and some of his Jamaican compatriots have been suspended for violations, Bolt has never been sanctioned or been declared to have tested positive for a banned substance.	& \faCheck	& \faClose	& \faCheck	\\
& & & & \\
15 &	The Utah man's mother, Laurie Holt, thanked Mr. Trump and the lawmakers for her son's safe return, adding: "I also want to say thank you to President Maduro for releasing Josh and letting him to come home."	& \faClose	& \faCheck	& \faCheck	\\
& & & & \\
16 &	They were fortunate to escape to America and to make good lives here, but we lost family in Kristallnacht.	& \faCheck	& \faCheck	& \faCheck	\\
& & & & \\
17 &	Historically, such consolidation (and price escalation) has enabled hospitals to offset higher expenses.	& \faCheck	& \faCheck	& \faCheck	\\
    \hline
    \end{tabular}
    \caption{Examples of annotated sentences from the CompRes dataset}
    \label{tab:examples}
\end{table*}

\section{Experiments}
\label{sec:experiments}

\begin{table*}
\centering
\begin{tabular}{ccccccccccccc}
\hline 

& \multicolumn{3}{c}{\textbf{SVM-BoW}} & \multicolumn{3}{c}{\textbf{RoBERTa}} & \multicolumn{3}{c}{\textbf{SVM-BoW-Aug}} & \multicolumn{3}{c}{\textbf{RoBERTa-Aug}} \\

& P & R & $F_1$ & P & R & $F_1$ & P & R & $F_1$ & P & R & $F_1$ \\
\hline 
Complication & 0.67 & 0.56 & 0.61 & \textbf{0.9} & 0.9 & 0.9 & 0.66 & 0.57 & 0.61 & 0.86 & \textbf{0.96} & \textbf{0.91} \\
Resolution & 0.48 & 0.34 & 0.4 & 0.63 & \textbf{0.66} & \textbf{0.64} & 0.5 & 0.38 & 0.43 & \textbf{0.76} & 0.45 & 0.57 \\
Success & 0.4 & 0.1 & 0.15 & 0.82 & 0.43 & 0.56 & 0.67 & 0.1 & 0.17 & \textbf{0.91} & \textbf{0.48} & \textbf{0.62} \\
\hline
Average & 0.52 & 0.33 & 0.39 & 0.78 & \textbf{0.66} & \textbf{0.7} & 0.61 & 0.35 & 0.4 & \textbf{0.84} & 0.63 & \textbf{0.7} \\
\hline
\end{tabular}
\caption{\label{tab:results} Precision (P), recall (R) and $F_1$ scores, evaluated on the test set, for the SVM model trained over bag-of-words features (\textbf{SVM-BoW}), the fine-tuned transformer-based language model (\textbf{RoBERTa}), the augmented SVM model trained over bag-of-words features (\textbf{SVM-BoW-Aug}) and the augmented fine-tuned transformer-based language model (\textbf{RoBERTa-Aug})}
\end{table*}

\subsection{Experimental Setup}

We randomly divided the news articles in the dataset into training, validation and test sets, while keeping the category distribution in the three sets as constant as possible; the statistics are given in Table \ref{tab:datasetdivision}. The training set was used to train the supervised model for the task; the validation set was used to select the best model during the training phase (further details are given in Sections \ref{subsec:baseline-model}), and the test set was used to evaluate the chosen model and produce the results reported in Section \ref{sec:results}.

\begin{table}
\centering
\begin{tabular}{lccc}
\hline & \textbf{\# Articles} & \textbf{\# Sentences} & \textbf{Ratio} \\ \hline
Training &	23 & 858 & 78\% \\
Validation & 3 & 115 & 11\% \\
Test & 3 & 126 & 11\% \\
\hline
\end{tabular}
\caption{\label{tab:datasetdivision} Training, validation \& test set statistics}
\end{table}

\subsection{Baseline Model}
\label{subsec:baseline-model}

For our baseline model, we used unigram counts (bag-of-words) as the feature representation. We first applied basic pre-processing to the texts: sentences were tokenized and lowercased, numbers were removed and contractions expanded. All the remaining terms were used as the features. We utilized a linear SVM classifier with the document-term matrix as input, and employed the one-vs-rest strategy for multilabel classification. 

The validation set was used to tune the $C$ hyperparameter for the SVM algorithm, via a random search on the interval $\left(0,1000\right]$, in order to choose the best model.

\subsection{Deep-Learning Model}

In addition to the baseline model, we experimented with a deep-learning model, fine-tuning a pre-trained language model for our multi-label classification task. We used the RoBERTa-base transformer \cite{liu2019roberta} as our base language model, utilizing the {\it transformers} python package \cite{wolf2019huggingfaces}. We appended a fully connected layer over the output of the language model, with three separate sigmoid outputs (one for each of the narrative categories), in order to fine-tune it to our task.

The entire deep model was fine-tuned for 5 epochs, and evaluated against the validation set after every epoch, as well as every 80 training steps. The checkpoint with the best performance (smallest loss) on the validation set was used to choose the best model.

\subsection{Data Augmentation}

Finally, we tested the effect of data augmentation in our setup; both models were re-trained with augmented training data, via back-translation. Back-translation involves translating training samples to another language and back to the primary language, thus increasing the size of the training set and potentially improving the generalization capacity of the model \cite{shleifer2019low}. For this purpose, we used Google Translate as the translation engine. Translation was performed to German and back to English, discarding translations that exactly match the original sentence. Following the augmentation, the training set size almost doubles in size, growing from 858 to 1683 samples. The validation set was used in the same way as in the pre-augmentation experiments.

\subsection{Results}
\label{sec:results}

We report our test results in Table \ref{tab:results}.

First, we observe that the deep models significantly outperformed the baseline models: an average $F_1$ score of 0.7 compared to 0.39/0.4, which represents an increase of 75\% in performance. The improvement is evident for every one of the narrative categories, but is particularly substantial for the \texttt{Success} category -- an $F_1$ score of 0.56 compared to 0.15, constituting an increase of 373\%. One plausible explanation we can offer has to do with the nature of our \texttt{Success} category; while the \texttt{Complication} and \texttt{Resolution} categories seem to be constrained by sets of generic terminologies, the definition of \texttt{Success} is more content-oriented, and thus highly sensitive to specific contexts. For example, linguistically speaking, the definition of the success of an athlete in never being tested positive for a banned substance (see sentence \#14 in Table \ref{tab:examples}) is very different from the definition of success in the cultural context of the art scene of a city (sentence \#10 in Table \ref{tab:examples}). 

Generally, the performance for each category appears to reflect the proportion of instances belonging to each category (see Table \ref{tab:datasetoverview}). This is most evident in the baseline models -- $F_1$ scores of 0.61, 0.4 and 0.15 in the SVM model, and $F_1$ scores of 0.61, 0.43 and 0.17 in the augmented SVM model for \texttt{Complication},  \texttt{Resolution} and \texttt{Success} (respectively). However, in the deep models this behavior seems to be less extreme; in the augmented RoBERTa model, the $F_1$ score for the \texttt{Success} category is higher by 0.05 compared to the \texttt{Resolution} category, despite being less frequent in the dataset. We also observe that the \texttt{Success} category consistently exhibit notably higher precision than recall, across all models, possibly due to the smaller number of samples encountered by the classifier during training. This is generally true for the \texttt{Resolution} category as well (except in the case of the RoBERTa model), though to a lesser extent. 

Interestingly, the data augmentation procedure does not seem to have any effect on model performance, both in the case of the baseline model (an increase of 0.01 in the average $F_1$ score) as well as the case of the deep model case (no change in the average $F_1$ score).

\section{Conclusion}
\label{sec:conclusion}

We introduced CompRes - the first dataset for narrative structure in news media. Motivated by the enormous social impact of news media and their role in creating and shaping of public opinion, we designed a new narrative structure annotation scheme which is better suited to informational text, specifically news articles. We accomplished that by adapting two elements from the theory introduced in \cite{labov1967narrative, labov2013language} -- \texttt{Complication} and \texttt{Resolution} -- and adding a new element, \texttt{Success}. This scheme was used to annotate a set of 29 articles, containing 1,099 sentences, which were collected from news and partisan websites.

We tested two supervised models on the newly created dataset, a linear SVM over bag-of-words baseline classifier and a fine-tuned pre-trained RoBERTa-base transformer, and performed an analysis of their performances with respect to the different narrative elements in our annotation scheme. Our preliminary results  -- an average $F_1$ score of up to 0.7 -- demonstrate the potential of supervised learning-methods in inferring the narrative information encoded into our scheme from raw news text.

We are currently engaged in an ongoing effort for improving the annotation quality of the dataset and increasing its size. In addition, we have several exciting directions for future work. First, we would like to explore incorporating additional elements from the narrative theory in \cite{labov2013language} to our annotation scheme; for example, we believe that the \textit{evaluation} element may be beneficiary in encoding additional information over existing elements in the context of news media, such as the severity of a \texttt{Complication} or the 'finality' of a \texttt{Resolution}. A related interesting option is to add completely new narrative elements specifically designed for informational texts and news articles, such as actor-based elements identifying entities which are related to one or more of the currently defined narrative categories; for instance, as mentioned in \ref{subsec:adaptation}, we may add indications for “villainous” protagonists, strongly associated with complications in the story and are expected to be instrumental to a successful resolution.

Another direction which we would like to explore includes enriching the scheme with clause-level annotation of the different narrative elements, effectively converting the task from multi-label classification to a sequence prediction one -- detecting the boundaries of the different narrative elements in the sentence. Alternatively, we could introduce additional layers of information which will encode more global narrative structures in the text, such as inter-sentence references between narratively-related elements (e.g., a \texttt{Resolution} referencing its inducing \texttt{Complication}), or even between narratively-related articles (e.g., different accounts of the same story).

\section*{Acknowledgements}
This research was partially supported by the Israel Science Foundation (grant No. 1400/14). We wish to thank Vered Porzycki and Avishai Green for their helpful comments on the CompRes codebook and for their careful tagging. 

\bibliography{anthology,acl2020}
\bibliographystyle{acl_natbib}

\end{document}